\documentclass[conference]{IEEEtran}
\IEEEoverridecommandlockouts

\usepackage{cite}
\usepackage{amsmath,amssymb,amsfonts}
\usepackage{algorithmic}
\usepackage{graphicx}
\usepackage{textcomp}
\usepackage{xcolor}
\usepackage{multirow} 
\usepackage{array}
\usepackage{threeparttable}
\def\BibTeX{{\rm B\kern-.05em{\sc i\kern-.025em b}\kern-.08em
    T\kern-.1667em\lower.7ex\hbox{E}\kern-.125emX}}

\begin{document}

\title{Enhanced Small Target Detection via Multi-Modal Fusion and Attention Mechanisms: A YOLOv5 Approach\\

}
\author{\IEEEauthorblockN{1\textsuperscript{st} Xiaoxiao Ma}
\IEEEauthorblockA{\textit{Northwestern Polytechnical University} \\
Xi'an, China \\
mxxnpu@mail.nwpu.edu.cn}
\and
\IEEEauthorblockN{2\textsuperscript{nd} Junxiong Tong}
\IEEEauthorblockA{\textit{University of Science and Technology of China} \\
Hefei, China \\
tongjunxiong@mail.ustc.edu.cn}

}

\maketitle

\begin{abstract}
With the rapid development of information technology, modern warfare increasingly relies on intelligence, making small target detection critical in military applications. The growing demand for efficient, real-time detection has created challenges in identifying small targets in complex environments due to interference. To address this, we propose a small target detection method based on multi-modal image fusion and attention mechanisms. This method leverages YOLOv5, integrating infrared and visible light data along with a convolutional attention module to enhance detection performance. The process begins with multi-modal dataset registration using feature point matching, ensuring accurate network training. By combining infrared and visible light features with attention mechanisms, the model improves detection accuracy and robustness. Experimental results on anti-UAV and Visdrone datasets demonstrate the effectiveness and practicality of our approach, achieving superior detection results for small and dim targets.
\end{abstract}


\section{Introduction}
The rapid advancement of information technology is reshaping modern warfare, transitioning it toward more intelligent and information-based forms. Modern military superiority no longer relies solely on advanced weaponry but also requires efficient information processing and decision-making systems. In this context, target detection and recognition have emerged as crucial technologies that significantly impact reconnaissance, surveillance, and precision strikes.

Among the various challenges in military operations, small target detection (such as micro-UAVs, stealth missiles, and small ground vehicles) holds particular importance\cite{b1}. These targets often possess characteristics such as small size and high mobility, making their accurate and real-time detection critical for modern warfare capabilities. However, real-world operational environments are highly complex and dynamic, presenting numerous interference factors such as complicated terrain, dynamic backgrounds, and various environmental noise. Accurately detecting small targets amidst these challenges has become a vital technological hurdle.

Traditional large-target detection algorithms have shown success in certain scenarios; however, when applied to small targets, their detection performance suffers due to limitations in resolution and feature extraction. This performance gap highlights the need for more sophisticated detection methodologies tailored specifically for small targets.

Multi-source information fusion is a promising technique that combines data from various sensors and information sources through advanced algorithms. The goal is to deliver a more accurate and comprehensive understanding than could be achieved with any single information source. In military applications, multi-source information fusion is particularly valuable. Different types of sensors, such as infrared sensors, radar, and visible-light cameras, each offer unique advantages and limitations depending on the scenario and target. For example, infrared sensors excel in night-time and adverse weather conditions but typically have lower resolution; radar can track targets over large distances but may struggle with reflection and noise in complex environments; visible-light cameras capture high-resolution imagery but are hindered by poor lighting conditions.

This research builds on the need for more robust small target detection systems by leveraging multi-source information fusion. Through the integration of infrared, radar, and visible-light data, this study seeks to enhance detection capabilities in challenging environments. Furthermore, this paper introduces innovative fusion algorithms designed to improve the accuracy and reliability of small target detection in real-world operational settings.

\section{Related Work}
In recent years, the problem of small object detection in images has attracted a lot of attention in both civil and military fields. Small target is the object occupying the least pixels in the image, and its low resolution and unobvious features are difficult to detect accurately\cite{b2}. To address these challenges, various approaches have been developed, each bringing unique benefits. Among them, the four most commonly used methods are multi-scale, super-resolution, attention mechanism and context-based strategy detection methods. The progress and examples of each method are described in detail below.

\subsection{Multi-Scale Detection Methods}
Multi-scale detection methods are essential in addressing the challenges posed by small objects, which are often difficult to detect due to their limited pixel representation. These methods leverage the extraction of features at different scales to enhance detection accuracy. Below, we outline two recent approaches that have made significant contributions in this area.

RetinaNet\cite{b3}, developed by Facebook AI Research, is a single-stage object detection algorithm that combines a Feature Pyramid Network (FPN) with Focal Loss to address class imbalance in detection tasks. The FPN enables the model to detect objects of various sizes by extracting multi-scale features, while Focal Loss focuses on hard-to-classify samples, improving detection of small objects obscured by larger elements. Its single-stage design simplifies and speeds up detection, enhancing small object performance.

Libra R-CNN\cite{b4}, introduced by Pang et al. in 2019, enhances multi-scale feature extraction for more consistent detection across object sizes. Its key feature, the Balanced Feature Pyramid (BFP), adaptively fuses multi-level features to improve small object representation in low-resolution maps. Libra R-CNN also employs IoU-balanced sampling and branch-balanced loss to tackle small object detection challenges, leading to better performance in complex environments.

\subsection{Super-Resolution Methods}
Super-resolution methods have emerged as a key approach to enhancing the detection of small objects by increasing the resolution of input images. These methods reconstruct high-resolution images from low-resolution inputs, allowing detection models to capture finer details. Below, we discuss two significant super-resolution techniques that have contributed to the advancement of small object detection.

FSRCNN (Fast Super-Resolution Convolutional Neural Network) \cite{b5} by Dong et al. is a fast and efficient method that builds on the SRCNN model. It innovates with a "down-scaling, convolution, and up-scaling" approach, performing convolutions in the low-resolution space to reduce computation while maintaining high-quality reconstruction. The final upscaling is done through a sub-pixel convolution layer, which enhances image detail and improves small object detection accuracy. For example, applying FSRCNN to the COCO dataset improved YOLOv3's small object detection performance by about 5

HRDNet (High-Resolution Detection Network)\cite{b6} by Liu et al. (2021) aims to boost detection performance while managing the computational load of high-resolution images. It combines a Multi-Depth Image Pyramid Network (MD-IPN) and a Multi-Scale Feature Pyramid Network (MS-FPN) to balance efficiency and object size variance. Using a deep backbone for low-resolution and a shallow backbone for high-resolution images, HRDNet captures detailed features for small objects, improving detection efficiency and effectiveness with high-resolution data.

\subsection{Attention Mechanism Methods}
Recent advancements in Transformer-based architectures have shown promise in improving small object detection through enhanced feature extraction mechanisms. Two notable approaches are Vision Transformer and Swin Transformer.

Vision Transformer (ViT)\cite{b7} applies the Transformer model, initially successful in natural language processing, to image processing. ViT divides an image into fixed-size patches, then embeds these patches linearly before feeding them into a Transformer encoder. This self-attention mechanism captures dependencies across different patches, enhancing the detection of small objects by focusing on relevant image features across the entire image. ViT's ability to model global attention helps improve performance in small object detection tasks.

Swin Transformer\cite{b8} , developed by Microsoft, builds on ViT by introducing a hierarchical structure and a sliding window mechanism. Unlike ViT, Swin Transformer uses window-based attention, which allows it to process image information more efficiently by focusing on local regions and shifting windows to capture global dependencies. This balance between local and global attention leads to higher computational efficiency and improved performance in small object detection, making it a robust approach for visual tasks.

\subsection{Context-Based Methods}
Context R-CNN\cite{b9} , proposed by Sara Beery et al. in 2020, is a context-aware object detection method designed to enhance the detection performance of small objects by leveraging surrounding contextual information. Traditional object detection struggles with small targets due to their limited features, but Context R-CNN improves recognition by integrating environmental context.

The method begins with feature extraction using a Convolutional Neural Network (CNN) to generate multi-layered feature maps. A Region Proposal Network (RPN) then identifies potential regions of interest. Context R-CNN further enhances detection by incorporating both short-term and long-term contextual information through attention mechanisms. It uses keyframe and non-keyframe features, combining information from multiple frames through two-stage Faster R-CNN detection and attention modules. This fusion of temporal and spatial context significantly boosts the accuracy of small object detection by providing a richer and more precise feature representation than conventional methods.

\section{Methodology}
his study presents a novel small target detection framework leveraging multi-modal image fusion combined with an attention mechanism, implemented on the YOLOv5 architecture. The proposed framework is designed to enhance the accuracy and robustness of detecting small and dim targets by integrating information from both infrared and visible light modalities, which each offer complementary advantages. Our approach also incorporates the Convolutional Block Attention Module (CBAM)\cite{b10} to further refine feature extraction, ensuring that the network focuses on the most relevant parts of the input data. The entire methodology is structured around four key stages: attention mechanism integration, multi-modal data fusion, image registration, and detection.

\subsection{CBAM Attention Module}
The CBAM module plays a critical role in refining feature representations by selectively emphasizing informative regions within the input feature maps. CBAM operates through two sequential sub-modules: the channel attention module and the spatial attention module.

The channel attention module focuses on refining the input feature map by emphasizing the importance of specific channels. Given an input feature map $F \in \mathbb{R}^{C \times H \times W}$, where $C$ represents the number of channels, and $H$ and $W$ denote the height and width of the feature map, channel attention is computed as:
\begin{equation}
M_c(F) = \sigma(\text{MLP}(\text{GAP}(F))),
\end{equation}

Here, $\text{GAP}(F)$ denotes the global average pooling operation, reducing each channel to a single value. This value is passed through a multi-layer perceptron (MLP) to compute the channel descriptor, followed by a sigmoid activation $\sigma$ to scale the descriptor between 0 and 1, representing the importance of each channel. The channel-attentive feature map is then combined with the original feature map via element-wise multiplication, which refines the feature representation based on channel importance.

Next, the spatial attention module operates on the channel-refined feature map, identifying the spatial locations within each feature map that contain the most critical information. A 2D convolution is applied across the feature map to produce a spatial attention map $M_s(F')$, which highlights the relevant spatial regions. The enhanced feature map is then obtained as:
\begin{equation}
F' = M_c(F) \cdot F + M_s(F') \cdot F.
\end{equation}
This combined attention mechanism allows the network to prioritize both the most informative channels and the most relevant spatial regions, enhancing the detection capabilities of the model, particularly for small objects that might otherwise be overlooked.

\subsection{Multi-modal Data Fusion}
One of the key innovations of our approach is the fusion of infrared and visible light data. Each modality provides unique benefits: infrared captures thermal information that is stable across varying lighting conditions, while visible light offers high-resolution texture and color details. Fusing these complementary modalities enhances the overall accuracy of the detection network, especially for small targets that are difficult to detect using a single modality.

Our fusion strategy is implemented at the feature map level. The infrared and visible light data are first processed independently through separate convolutional streams, each producing corresponding feature maps $F_{\text{IR}}$ and $F_{\text{VIS}}$. The feature maps are then combined through a weighted summation, where the weight parameter $\alpha
$ dynamically adjusts based on the relative importance of each modality:
\begin{equation}
F_{\text{fused}} = \alpha \cdot F_{\text{IR}} + (1 - \alpha) \cdot F_{\text{VIS}}.
\end{equation}
The weight $\alpha$ is determined based on the statistical properties of the feature maps, ensuring that the network optimally fuses information from both modalities for improved target detection. The fused feature maps contain a richer set of information, allowing the network to exploit the strengths of both data sources in challenging environments.

\subsection{Image Registration}
An essential preprocessing step in our framework is image registration, which ensures that the infrared and visible light images are spatially aligned before fusion. Given the different imaging properties of the two modalities, accurate alignment is crucial for the subsequent fusion and detection stages.

We perform image registration using a homography-based transformation. The goal is to align corresponding points between the infrared and visible images by minimizing the reprojection error. Let $p_i$ and $p_i'$ be the coordinates of corresponding points in the visible and infrared images, respectively. The transformation matrix $H$ is computed by minimizing the sum of squared distances between the projected and actual points:
\begin{equation}
H = \arg\min_H \sum_i \| \mathbf{p}_i' - H \mathbf{p}_i \|^2.
\end{equation}
This transformation aligns the images, ensuring that the fused feature maps correctly correspond to the same physical locations in both modalities. This accurate alignment is particularly important in small target detection, where even minor misalignments can significantly degrade performance.

\subsection{Final Detection Framework}
Once the feature maps are fused and attention has been applied, the combined feature maps are passed through the detection head of the YOLOv5 architecture. YOLOv5 is an efficient, single-stage object detector that directly predicts bounding boxes, class scores, and objectness scores for each detected object.

The detection head in YOLOv5 is composed of several convolutional layers that reduce the high-dimensional fused feature maps to the required output dimensions. Specifically, the network predicts three outputs for each bounding box: the localization (bounding box coordinates), objectness score, and class probabilities. These predictions are trained using a multi-task loss function that combines losses for each of these tasks:
\begin{equation}
\mathcal{L} = \lambda_{\text{loc}} \cdot \mathcal{L}_{\text{loc}} + \lambda_{\text{conf}} \cdot \mathcal{L}_{\text{conf}} + \lambda_{\text{cls}} \cdot \mathcal{L}_{\text{cls}},
\end{equation}
Here, $\mathcal{L}_{\text{loc}}$ represents the localization loss, which measures the accuracy of the predicted bounding box coordinates. $\mathcal{L}_{\text{conf}}$ is the confidence loss, which measures how well the model can differentiate between background and object regions, while $\mathcal{L}_{\text{cls}}$ is the classification loss, measuring the accuracy of the predicted class labels. The weights $\lambda_{\text{loc}}$, $\lambda_{\text{conf}}$ and $\lambda_{\text{cls}}$ are hyperparameters that balance the contributions of each loss term during training.

This multi-task learning framework enables the network to simultaneously optimize for accurate localization, object classification, and confidence estimation, ultimately leading to improved performance on small target detection tasks.

\subsection{Training and Evaluation}
To validate the effectiveness of our approach, we train and evaluate the model on two datasets: the anti-UAV dataset and the Visdrone dataset. Both datasets contain small and dim objects in challenging environments, providing a robust testing ground for the proposed method. The model is trained using standard stochastic gradient descent (SGD) with a learning rate scheduler to adaptively adjust the learning rate during training.

The evaluation metrics used to assess the model include precision, recall, and the F1 score, which provide a comprehensive view of the model's performance in terms of both accuracy and robustness. Additionally, we evaluate the detection speed to ensure that the model meets real-time performance requirements, a critical aspect of many military applications.

\section{Experiments and Results}

\subsection{Experimental Settings}
The experiments were conducted on a workstation equipped with an Intel® Core™ i7-14700KF CPU, 16GB of RAM, and an NVIDIA GeForce RTX 4060 Ti GPU with 8GB of VRAM. The system was running on Windows 11 with CUDA version 12.5. The main software platforms utilized included PyCharm 2023.1, Anaconda3, PyTorch 2.2.1, ClearML 1.16.1, Jupyter 1.0.0, OpenCV-Python 4.9.0.80, Python 3.11.8, and Weights \& Biases (wandb) 0.12.10.

\subsection{Datasets}
Two main datasets are used in this study: the anti-UAV dataset\cite{b11} and the VisDrone2019\cite{b12} dataset. The anti-UAV dataset mainly consists of videos used for small object detection in anti-UAV scenarios, while the VisDrone2019 dataset is widely used for object detection tasks involving small aerial objects. These datasets comprehensively cover a variety of scenarios related to small object detection and were chosen because of their relevance to the problem at hand.

For this study, the anti-UAV video dataset is processed to create a registered multimodal image dataset. The preprocessing step consists of extracting frames from the original video and labeling them accordingly. The extracted frames include infrared and visible images, which are unaligned in their original form. The feature point matching algorithm is used to register their images, which guarantees the accurate registration of infrared images with visible images, and this alignment process is essential to prepare a coherent dataset suitable for training and evaluating the proposed model. The infrared and visible images after registration are saved as two new datasets, IR and VIS datasets. In the following, experiments will be carried out on IR and VIS datasets respectively. The results are compared with those obtained after the fusion of IR and VIS.

\subsection{Results on Anti-UAV Dataset}

In contrast experiments with different data sets, the hyperparameters of model training are set to fixed values in order to control variables. Set the training rounds to 50, batch size to 16, image size parameter to 640, pre-training weight and model profile to yolov5l.

The original model was used to train the infrared and visible image data sets respectively, and then the model with the attention module was used to train the infrared and visible light data sets respectively, and their $F_1$ curve peak and mAP@0.5 were observed. The comparison results are shown in Table 1. As can be seen from Table 1, when the other training conditions are fixed, the mAP@0.5 trained on the visible light dataset is 0.781 when the attention module is not added, and it rises to 0.874 after the attention module is added, which is a significant improvement of 11.9\%. The mAP@0.5 trained on the infrared dataset is 0.898, which rises to 0.958 after adding the attention module, also obtaining a 6.7\%improvement, which shows that the addition of the attention module has a significant effect on the increase of the average detection accuracy.

For the $F_1$ curve, when the other training conditions were fixed, the highest $F_1$ value of 0.77 was obtained when the confidence was 0.339 when the attention module was trained on the visible light dataset, and the highest value of 0.81 was obtained when the confidence was 0.471 after the attention module was added. After training on the infrared dataset, the highest $F_1$ value of 0.84 is obtained when the confidence is 0.353, and the highest $F_1$ value of 0.96 is obtained when the confidence is 0.660 after adding the attention module. The highest $F_1$ value and the confidence when the highest value are improved to different degrees on both the infrared and visible light datasets. It shows that the performance is more balanced after adding the CBAM module, which not only effectively reduces the missed detection (improves the recall rate), but also reduces the false detection (improves the precision rate).

When the CBAM module and fusionnet are added at the same time, mAP@0.5 trained on the anti-uav dataset reaches 0.977, which is 8.8\% higher than the original highest 0.898. The F1 value also reaches a peak of 0.99 at 0.431, which is higher than the original highest of 0.96. The results show that the new network architecture has reached the current optimal level in detection accuracy and balance.

\begin{table}[htbp]
\centering
\begin{threeparttable}
\caption{Performance of different models on the anti-uav dataset}
\centering
\begin{tabular}{|c|c|c|c|}
  \hline
             & $F_1$ curve (peak)/Confidence & mAP@0.5 \\ \hline
  YOLOv5(IR) & 0.84 / 0.353 & 0.898\\ \hline
  YOLOv5+CBAM(IR) & 0.96 / 0.660 & 0.958 \\ \hline
  YOLOv5(VIS) & 0.77 / 0.339 & 0.781 \\ \hline
  YOLOv5+CBAM(VIS) & 0.81 / 0.471 & 0.874 \\ \hline
  YOLOv5+CBAM+fusionnet & \textbf{0.99 / 0.431} & \textbf{0.977} \\ \hline
\end{tabular}
\begin{tablenotes}
\footnotesize
\item[IR] IR refers to the results trained on the infrared dataset.
\item[VIS] VIS refers to the results trained on the visible light dataset.
\end{tablenotes}
\end{threeparttable}
\end{table}

In summary, after the fusion of infrared and visible images, the comprehensive accuracy and recall rate of the model are better than the indicators of visible or infrared image detection alone, indicating that the fusion of dual-modality can indeed improve the performance of the model in small target detection.

\subsection{Results on Visdrone Dataset}

As shown in Fig 2, in the comparison experiment of visdrone dataset, when other training conditions are fixed, the average $F_1$ value of all categories reaches the maximum value of 0.38 when the confidence is 0.193 when the attention module is not added, and this value reaches the maximum value of 0.43 when the confidence is 0.206 when the attention module is added. Among them, the $F_1$ value of tricycle, awning-tricycle, bus, motor and other categories has been greatly improved, indicating that after adding the attention module, the balance of the model training on the Visdrone dataset has been improved.

\begin{figure}[htbp]
\centering
\includegraphics[width=0.35\textwidth]{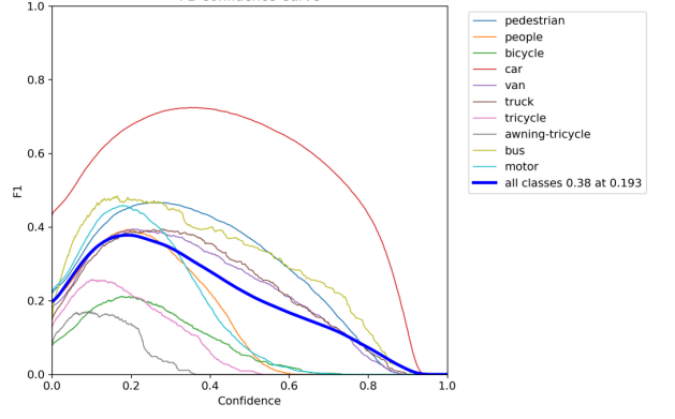} 
\includegraphics[width=0.34\textwidth]{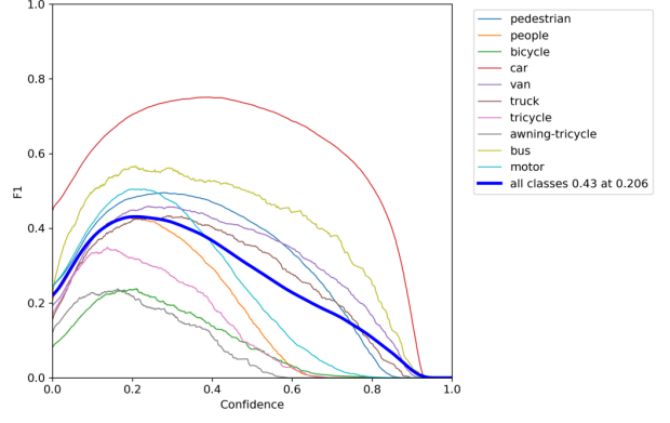} 
\caption{The top/bottom figures show the $F_1$ curves of Visdrone dataset trained on the original model/the model with attention module, respectively.}
\label{fig}
\end{figure}

As can be seen from the $PR$ curve of Fig 3, when the other training conditions are fixed, the mAP@0.5 of each category is 0.337 when the $IoU$ of each category is 0.5 when the attention module is not added, and the value rises to 0.395 when the attention module is added, which is a significant improvement. The biggest improvement is in the bus class, where the accuracy is increased from 0.410 to 0.536. The obvious rise of this index indicates that the addition of the attention module can significantly improve the accuracy of small target detection in the Visdrone dataset.

\begin{figure}[htbp]
\centering
\includegraphics[width=0.33\textwidth]{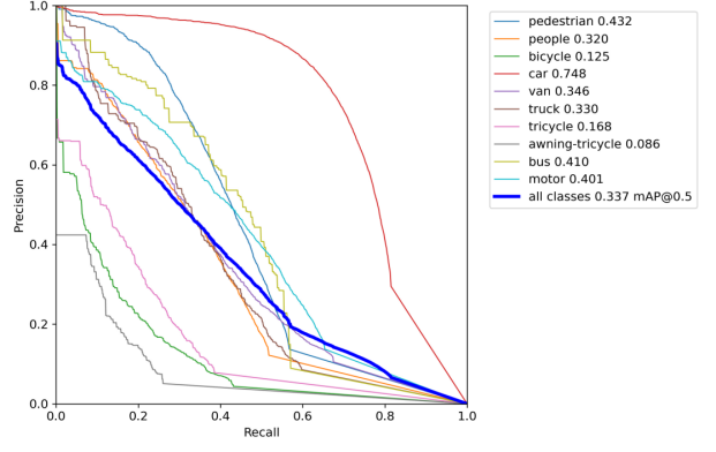} 
\includegraphics[width=0.33\textwidth]{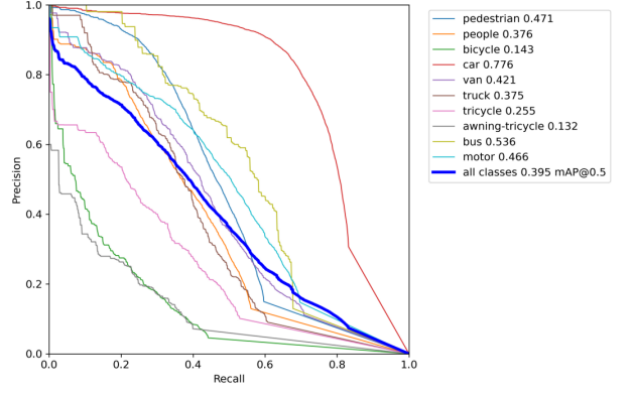} 
\caption{The top/bottom figures show the $PR$ curves of Visdrone dataset trained on the original model/the model with attention module, respectively.}
\label{fig}
\end{figure}

\section{Conclusion}

In this paper, we propose a new method to enhance the ability of small object detection, which fuses infrared and visible dual-modal images and adds an attention mechanism. Experimental results on the anti-UAV and Visdrone datasets show that the proposed model is superior to the existing methods in terms of detection accuracy and robustness.

The main contribution of our work is to add the CBAM attention module to improve the model's ability to pay attention to the key features of small targets, and to supplement the shortcomings of easy to miss and low detection accuracy in single modal detection by fusing infrared and visible dual-modal images, which improves the robustness of the model and better balances the accuracy and recall rate of the model.

In summary, the proposed model represents an important step in the research field of small object detection, and provides an adaptive and effective solution to improve the ability of small object detection using sensor data from different modalities.



\end{document}